\documentclass[APA,STIX2COL]{WileyNJD-v2}

\usepackage{threeparttable}
\usepackage{tabularx}
\usepackage{silence}
\usepackage[switch]{lineno}

\articletype{PREPRINT ARTICLE}%

\received{26 April 2016}
\revised{6 June 2016}
\accepted{6 June 2016}

\begin{document}

\title{Semi-Supervised Pipe Video Temporal Defect Interval Localization}

\author[1]{Zhu Huang}

\author[1]{Gang Pan*}
\author[2]{Chao Kang}
\author[3]{YaoZhi Lv}

\authormark{Zhu Huang \textsc{et al}}

\address[1]{
  \orgdiv{College of Intelligence and Computing},
  \orgname{Tianjin University},
  \orgaddress{
    \state{Tianjin},
    \country{China}
  }
}

\address[2]{
  \orgname{University of Alberta},
  \orgaddress{
    \state{Edmonton, Alberta},
    \country{Canada}
  }
}

\address[3]{
  \orgdiv{Key laboratory of Infrastructure Durability},
  \orgname{Tianjin Municipal Engineering Design and Research Institute},
  \orgaddress{
    \state{Tianjin},
    \country{China}
  }
}

\corres{*Gang Pan, College of Intelligence and Computing, Tianjin University, No. 135 Yaguan Road, Tianjin 300350, China.\\\email{pangang@tju.edu.cn}}

% \presentaddress{This is sample for present address text this is sample for present address text}

\abstract[Summary]{
  In sewer pipe Closed-Circuit Television (CCTV) inspection, accurate temporal defect localization is essential for effective defect classification, detection, segmentation and quantification. Industry standards typically do not require time-interval annotations, even though they are more informative than time-point annotations for defect localization, resulting in additional annotation costs when fully supervised methods are used. Additionally, differences in scene types and camera motion patterns between pipe inspections and Temporal Action Localization (TAL) hinder the effective transfer of point-supervised TAL methods. Therefore, this study introduces a Semi-supervised multi-Prototype-based method incorporating visual Odometry for enhanced attention guidance (PipeSPO). PipeSPO fully leverages unlabeled data through unsupervised pretext tasks and utilizes time-point annotated data with a weakly supervised multi-prototype-based method, relying on visual odometry features to capture camera pose information. Experiments on real-world datasets demonstrate that PipeSPO achieves 41.89\% average precision across Intersection over Union (IoU) thresholds of 0.1-0.7, improving by 8.14\% over current state-of-the-art methods.
}


\maketitle

\section{Introduction}

Regular inspection and maintenance are crucial for complex sewer pipe systems~\citep{us_pipe_length_2010,review_water_2022}. Currently, the primary method for detecting sewer pipe defects is the Closed-Circuit Television (CCTV) inspection system. Technicians use remotely controlled robots equipped with cameras to traverse the interior of the pipes and capture video, which is then manually assessed to locate defects. However, manual assessment suffers from inefficiencies, lack of objectivity, and susceptibility to oversight. Automated algorithms have been developed to detect defects in collected videos~\citep{ma_TransformeroptimizedGenerationDetection_2023,wang_ConstructionMaintenanceUrban_2022,yin_AutomationSewerPipe_2021,wang_AutomatedSewerPipe_2021,xiao_VisionbasedMethodAutomatic_2021}. However, existing methods still fail to reach the industry requirements in terms of precision and recall. Consequently, the current focus of computer technology should be on improving the efficiency of inspecting process by enhancing assistance and support for technicians rather than replacing them.

During the assessment stage, technicians must first identify the approximate time-intervals when pipe defects encounter. They then examine the footage frame by frame to select key frames that reasonably indicate the actual condition of the defects for inclusion in the inspection report. Subsequently, they perform further tasks such as defect classification~\citep{li_SewerPipeDefect_2021}, detection~\citep{li_AttentionguidedMultiscaleNeural_2023,dang2022defecttr,li_SewerPipeDefect_2021}, segmentation~\citep{maattention2024,fang_SewerDefectInstance_2022,pan2020automatic,wang2020unified}, and quantification~\citep{zhou2022automatic,xiong2022intelligent}. Therefore, the Pipe CCTV Video Temporal Defect Interval Localization (CTDIL) task is a prerequisite for subsequent operations. However, since pipe defects only appear in short time-intervals—comprising approximately 15\% of the footage in this study's dataset—utilizing algorithms to assist technicians in the CTDIL task is crucial for reducing the costs of assessing CCTV inspected videos.~\citep{cost_optimization_2023}.

In practice, sewer pipe CCTV inspection reports typically do not require the annotation of defect time-intervals, only a key frame indicating a defect, which limits the application of fully supervised learning methods from Temporal Action Localization (TAL) tasks to the CTDIL task. Existing methods that utilize computer vision techniques for assisting pipe defect detection are primarily based on frame-by-frame image classification, supplemented by post-processing methods such as Kalman filtering and metric learning. Although these approaches can localize defects, they do not directly analyze the inspection video and thus fail to effectively leverage the temporal information inherent in the video~\citep{ma_TransformeroptimizedGenerationDetection_2023,yin_AutomationSewerPipe_2021,wang_AutomatedSewerPipe_2021,xiao_VisionbasedMethodAutomatic_2021,YIN2020,li2019sewer,HAWARI2018}.

\begin{figure}
  \centering
  \includegraphics[width=\linewidth]{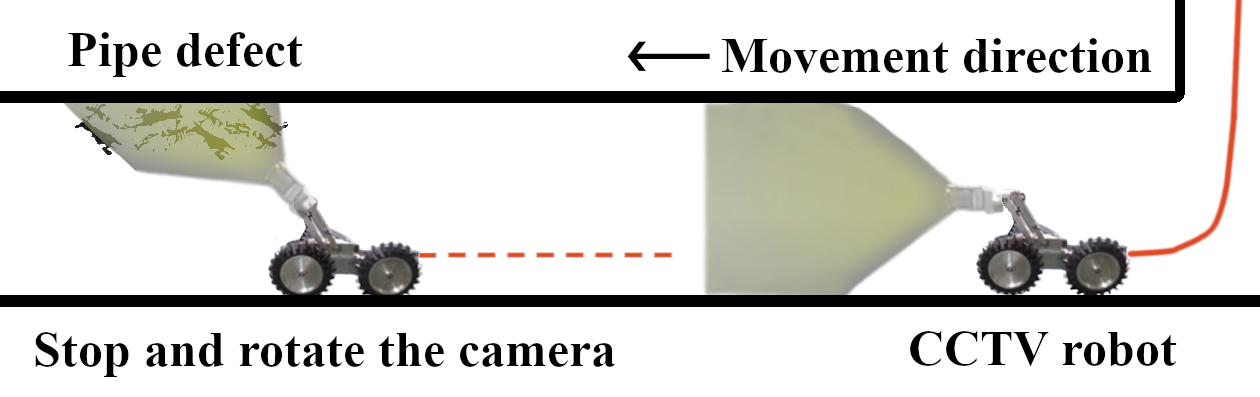}
  \caption{Illustration of the CCTV robot used in sewer pipe inspection. When no pipe defect is observed, the robot moves in a straight line with the camera facing forward, and upon observing a pipe defect, it typically stops and rotates the camera to align with the defect, indicating a strong relationship between camera pose changes and the occurrence of pipe defects. Leveraging this prior knowledge, visual odometry features can provide attention guidance to the model, thus enhancing its ability to locate pipe defects.}
  \label{fig:CCTV_VO}
\end{figure}

Although point-supervision TAL algorithms~\citep{zhang2023hr,LACP_2021,ma2020sf} can be directly used, there are significant differences between pipe scenes and human activity scenes, resulting in weaker feature representation capabilities when using general feature extractors, which further hinders the distinction between defect features and background features.~\citep{pipetr_2024} Moreover, the camera motion patterns in pipe inspection videos are entirely different from those in human activity videos. As shown in Figure~\ref{fig:CCTV_VO}, in sewer pipes inspection, robots carrying cameras move in a straight line and usually stop to rotate the camera for a few seconds to capture clearer images upon observing a defect. This provides excellent clues, but current point-supervised TAL algorithms cannot utilize the detailed information. These factors collectively lead to poor performance when directly using TAL methods~\citep{videopipe_2022}. Lastly, due to management negligence and other reasons, there are a number of unannotated pipe inspection videos in archived data, which cannot be utilized by supervised learning methods.

To address aforementioned issues, this study introduces a sewer pipe-focused video-based semi-supervised multi-prototype-based model utilizing visual odometry technology to capture camera pose information (PipeSPO). First, the feature extractor's representation capability is enhanced through an unsupervised pretext task, improving the performance of a pre-trained feature extractor in pipe scenes. Then, weakly supervised training based on multiple prototypes is employed to extract defect features across videos. By using a clustering method to obtain multiple prototype features and incorporating a contrastive learning loss function, the issue of weak feature extraction capability is further mitigated. Finally, a pre-trained visual odometry feature is used to train an attention-guided model, along with a gating mechanism, to further enhance the model's performance. The PipeSPO model capitalizes on the unique characteristics of pipe CCTV inspection videos, introducing a video-based semi-supervised approach to the CTDIL task, which significantly enhances performance. This has crucial practical implications for the maintenance and repair of urban infrastructure.

\section{Related Work}

\subsection{Temporal defect localization}

\subsubsection{Image-based methods}

The pipe video temporal defect localization algorithms are primarily based on image classification. These algorithms typically classify each frame of the inspection video, combined with additional processing steps such as manual post-processing and independent tracking models, to obtain the assessment results. For example,~\cite{wang_AutomatedSewerPipe_2021} uses the Faster R-CNN network for object detection frame by frame, combined with metric learning methods to extract distinguishing features of defects, and then uses Kalman filtering for instance tracking. ~\cite{ma_TransformeroptimizedGenerationDetection_2023} employs two transformer-based networks. First, the video frames are sent to the detection network; upon detecting a pipe defect, the frame is then sent along with the previously detected defect-containing frame to the tracking network to assess if they belong to the same defect, thus localizing the pipe defect. In the VideoPipe competition held by~\cite{videopipe_2022}, the method using frame-by-frame classification combined with a peak-seeking algorithm performed better. The inability to directly utilize temporal information limits the application of these methods.

\subsubsection{Video-based methods}

Although there are many methods for directly analyzing videos in the TAL~\citep{liu2024end, shi2023tridet, actionformer_2022}, video-based methods are not common for temporal localization tasks in pipe inspection videos. ~\cite{pipetr_2024} proposed a fully supervised video-based method using the transformer's global analyzing and parallel computing capabilities. This approach integrates multi-source features and directly analyzes the video, enabling the output of defect time-points without the need for complex post-processing steps, achieving promising results. However, due to the lack of time-interval level annotations, this method cannot be directly applied to the CTDIL task with time-point level annotations.

\subsection{Point-supervised temporal action localization}

Due to the lack of time-interval level annotations, fully supervised algorithms in mainstream TAL tasks cannot be utilized. However, a few point-supervised TAL algorithms has been proposed. ~\cite{ma2020sf} introduced a point-supervised TAL task and the SF-Net model, which mines potential background and action frames from unannotated frames and uses these pseudo-labels as training data. ~\cite{LACP_2021} improved upon SF-Net by generating higher quality pseudo-labels using a greedy algorithm based on action completeness, thus enhancing model performance. ~\cite{zhang2023hr} employed a two-stage method: in the first stage, prototype features of each action class from the training dataset are stored and dynamically updated during training. An attention mechanism is used to enable the model to learn action features across videos, generating high-quality pseudo-labels. In the second stage, these pseudo-labels are matched with true annotations to further fine-tune the final predictions.

However, due to significant differences in video scenes, feature extractors pre-trained in regular scenarios perform poorly in pipe scenes, further complicating the distinction between defect features and background features. Additionally, the camera movement patterns in these two types of scenes differ significantly. Consequently, the methods used in TAL tasks are poorly aligned with the requirements of the CTDIL task.

\section{PipeSPO}

\subsection{Dataset}
\begin{figure}
  \centering
  \includegraphics[width=\linewidth]{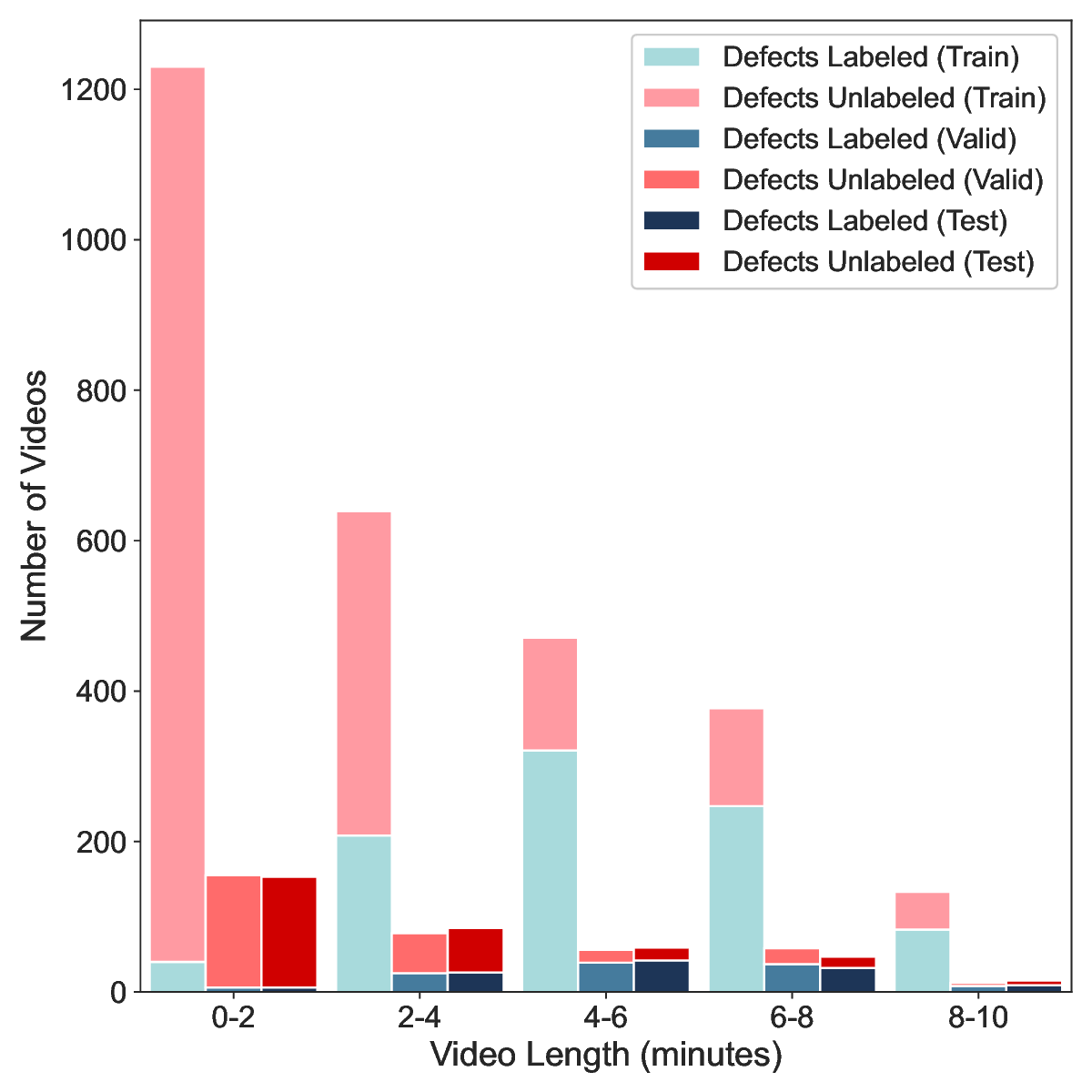}
  \caption{Distribution of video lengths. Only the test set contains time-interval level annotations, while the rest contains only time-point level annotations.}
  \label{fig:video_length_distribution}
\end{figure}

The dataset is extracted from pipe inspection reports provided by the Tianjin Municipal Engineering Design and Research Institute and datasets from the VideoPipe competition~\citep{videopipe_2022}, which contain only time-point level annotations. The test set annotations were manually expanded to time-interval level for model performance evaluation. Unlabeled data was used exclusively in the unsupervised pretext task.

In this study, a hash algorithm was used to remove potentially duplicate video data, followed by preprocessing using the OpenCV library~\citep{opencv_library} as follows: resizing the frames to 224 pixels in height and width; downsampling the videos to 3 frames per second (FPS); and segmenting videos longer than 10 minutes. The data were then divided into annotated and unannotated parts and split into training, validation, and test sets in an 8:1:1 ratio. For annotated video data, the split considered the distribution of video durations, the ratio of videos with and without pipe defects, and the distribution of defect occurrences in defect-containing videos. For unannotated video data, the split aimed to maintain a similar distribution of video durations across all parts.

The final dataset contains approximately 3500 videos, with a total duration of 185 hours and an average video duration of 3 minutes. Annotated videos account for about 100 hours, while unannotated videos account for about 85 hours. Among the annotated videos, those containing pipe defects span approximately 46 hours, with about 2700 defect-containing videos. On average, each defect-containing video includes 5 pipe defects, with approximately 49\% of these videos containing only one defect. The duration distribution of the dataset is shown in Figure~\ref{fig:video_length_distribution}, where the duration distribution of annotated videos is relatively uniform, while many unannotated videos are shorter than 2 minutes.

\subsection{Problem setting}

Similar to the Temporal Action Localization (TAL) task, the CTDIL task focuses on identifying defect time-intervals in pipe inspection videos. Given a set of videos \( \mathcal{V} = \{V_1, V_2, \ldots, V_n\} \), each \( V_i \) is associated with annotations indicating defect time-points \( \mathcal{D}_i = \{t_1, t_2, \ldots, t_m\} \). In cases where \( V_i \) has no defects, \( \mathcal{D}_i = \emptyset \) (an empty set). In the test set only, there are annotations indicating defect time-intervals \( \mathcal{D}_i' = \{[t_{1,\text{start}}, t_{1,\text{end}}], [t_{2,\text{start}}, t_{2,\text{end}}], \ldots, [t_{m,\text{start}}, t_{m,\text{end}}]\} \), which are used for network evaluation. The objective is to accurately determine the set \( \mathcal{D}_i \) for each \( V_i \), highlighting the occurrence of pipe defects during specific time-intervals.

\subsection{Feature extraction}\label{sec:feature_extraction}

Given the extended duration of the videos, direct usage as input to the network requires a significant amount of GPU memory, thus necessitating feature extraction. The PipeSPO model utilizes three methods for feature extraction: a static feature extractor, a dynamic information extractor, and a visual odometry extractor.

\subsubsection{Static and dynamic information feature}

Static and dynamic features are mainly used in video processing~\citep{kinetics_2017,two_stream_2014}. The static and dynamic information feature extractors utilized in this study are based on methods proposed by~\cite{pipetr_2024}. Specifically, the static feature extractor utilizes a custom-trained pipe defect image classification network. For the dynamic information feature extractor, a ResNet-151 model~\citep{resnet_2016} pre-trained on ImageNet~\citep{imagenet_2009} is used to extract the LLH sub-bands of pipe videos, which are decomposed by a 3D discrete wavelet transform (3D-DWT), thereby obtaining dynamic features from the video frames.

\subsubsection{Visual odometry feature}

Since the variations in camera position and pose in pipe inspection videos are strongly correlated with the occurrence of pipe defects, using a pre-trained visual odometry network to extract features can aid in locating these defects. Precise monocular visual odometry relies on camera intrinsic parameters~\citep{klenk2024deep,teed2023deep,alvarez2023monocular,franccani2022dense}. Through transformation matrices, points in the camera coordinate system can be mapped into the world coordinate system to obtain the camera's pose. However, due to the unavailability of intrinsic parameters for the cameras used to capture the dataset's videos, this study employs the pre-trained visual odometry network DeepVO~\citep{wang2017deepvo}, which does not require camera intrinsics. By removing the network's final fully connected layer, feature representations of the camera pose information from the videos are extracted.

Due to the generalization capability of deep learning networks, useful features can still be extracted across different datasets even without specific camera intrinsics. In the pipe defect localization algorithm, the focus is on the relative movement of the camera, i.e., forward/backward motion and pose changes, rather than absolute position or precise distances. Visual odometry networks excel at capturing and understanding this relative movement from sequential image frames, which reduces the dependency on camera intrinsics. The aim of the camera pose feature extractor is to identify changes in camera pose or specific time-intervals in video sequences when the camera position remains unchanged. Visual odometry features provide crucial cues about camera motion and viewpoint changes, thereby guiding another network to focus on these critical moments. Therefore, a guiding module based on visual odometry features is designed. Even if the accuracy of these features is not very high, as long as they reflect the general motion trend and pose changes of the camera, they are sufficient to guide another network's focus.

\subsection{Model architecture}\label{sec:pipespo_architecture}

\begin{figure*}[t]
  \centering
  \includegraphics[width=\linewidth]{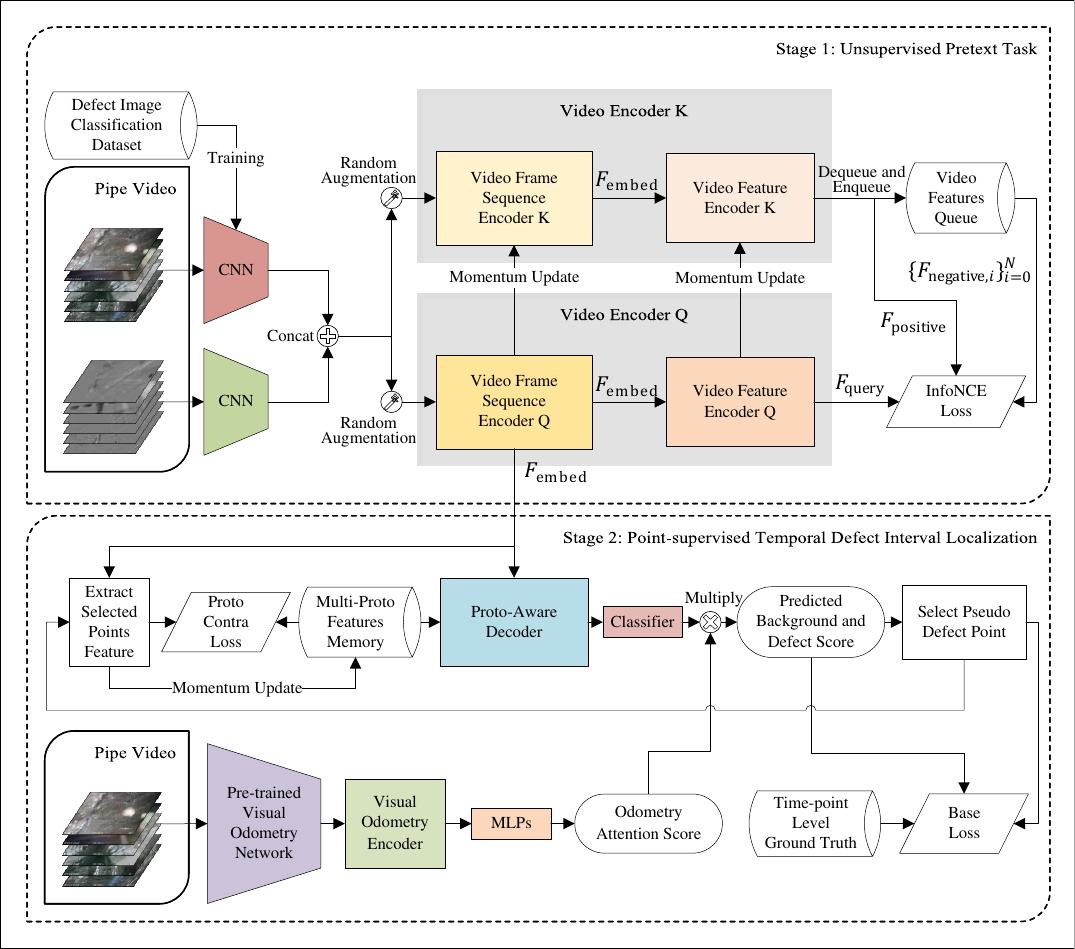}
  \caption{PipeSPO architecture. The network modules with the same color share weights. PipeSPO consists of two stages: the first stage is an unsupervised pretext task that trains a video frame sequence encoder using unlabeled videos; the second stage is a semi-supervised temporal defect interval localization, utilizing the clustering-based multi-prototype memory and prototype perception module, and incorporating camera pose information to guide the network. For a detailed introduction, see section~\ref{sec:pipespo_architecture}.}
  \label{fig:PipeSPO_Architecture}
\end{figure*}

The overall framework of PipeSPO is illustrated in Figure~\ref{fig:PipeSPO_Architecture}. The entire algorithm consists of two stages: the first stage is an unsupervised pretext task that trains a video frame sequence encoder using annotated and unannotated videos. The second stage is a point-supervised temporal defect interval localization, utilizing the clustering-based prototype memory and prototype perception module, and incorporating camera pose information to guide the network.

\subsubsection{Unsupervised pretext task}

In the unsupervised pretext task phase, PipeSPO was trained using all inspection videos in the training set, including both labeled and unlabeled videos. Inspired by the MoCo method~\citep{he2020momentum}, the network includes two identical deep video encoders $VideoEncoder_Q$ and $VideoEncoder_K$ that map the raw input feature sequence of videos to video features, which do not share weights. During training, both encoders are initialized with the same random weights. $VideoEncoder_K$ does not record gradients and uses momentum updating, where its weights are slowly updated using the weights of $VideoEncoder_Q$.

In unsupervised contrastive learning, due to the lack of labeled data, features are captured through the data itself. Each video is treated as an independent label to compute the loss, which requires a very large batch size; otherwise, the network may fail to learn meaningful features. If end-to-end methods are used, recording gradients for both encoders demands significant GPU memory. By not updating weights of $VideoEncoder_K$, GPU memory requirements are reduced, but the difference between positive and negative sample encoders increases, reducing negative sample consistency. Momentum updating maintains negative sample consistency without significantly increasing GPU memory requirements, thereby preserving a large negative sample feature set.

At the start of training, a video negative sample feature queue is randomly initialized and dynamically updated during training. Each batch contains query samples and their corresponding positive samples, obtained from $VideoEncoder_Q$ and $VideoEncoder_K$, respectively. The remaining samples in the negative sample queue are treated as negative samples. After computing the contrastive loss, the features from $VideoEncoder_K$ are added to the negative sample queue, replacing the earliest features. This ensures that the negative sample queue is continuously updated, enhancing network performance.

Two feature extractors mentioned in section~\ref{sec:feature_extraction} are utilized as raw features: a static feature extractor based on a pipe defect image classification network and a dynamic information extractor based on 3D-DWT. Considering a video batch of size $B$, where the longest video has $T_{max}$ frames, video resolution is $H \times W$, and each frame has $C$ channels. Initially, video data undergo sequence padding to unify their lengths to $T_{max}$, resulting in the video tensor:
\begin{equation}
  V_{input} = [B, T_{max}, H, W, C]
\end{equation}
Subsequently, the video sequence is fed into the two feature extractors, each with feature dimension $D$. The feature tensors obtained are:
\begin{equation}
  F_{cls}, F_{dwt} = [B, T_{max}, D]
\end{equation}
These tensors are concatenated to form the raw input feature:
\begin{equation}
  F_{input} = [B, T_{max}, 2D]
\end{equation}
Next, the raw input features are duplicated and subjected to different random augmentations, such as randomly masking some frames, shuffling a small region of frames, and mixing noise with the original features. These augmented features are combined with absolute position encoding and input into the Encoder module to obtain video embedding query features:
\begin{equation}
  F_{query} = VideoEncoder_Q(RandomAug(F_{input}))
\end{equation}
Another set of augmented raw input features is used as positive samples:
\begin{equation}
  F_{positive} = VideoEncoder_K(RandomAug(F_{input}))
\end{equation}
The video sequence feature encoder uses a Transformer Encoder architecture~\citep{transformer_2017} to capture global information from video sequences and re-embed the raw features. The video-level feature encoder uses a multi-layer cross-attention structure to convert variable-length video sequences into fixed-length features, which are then averaged through a pooling layer. This method retains temporal information better than direct average pooling.

$F_{query}$ and $F_{positive}$ should be similar since the same sample data use different augmentations, preserving the video semantics. In the CTDIL task, this implies that the temporal location information of pipe defects in the inspection video should remain consistent. Negative samples are features from the negative sample queue, obtained using the momentum-updated $VideoEncoder_K$. This design forces the network to learn temporal information from video sequences, enhancing its time-interval localization capability and improving feature embedding.

\subsubsection{Point-supervised temporal defect interval localization}

In the point-supervised temporal defect interval localization phase, PipeSPO was trained only based on labeled videos. In this phase, PipeSPO uses a clustering-based approach to cluster defect and background time-point features in the dataset, obtaining prototype features of defect and background time-points. These prototype features are updated with momentum during training, providing additional supervision signals through prototype contrastive loss. Additionally, a pre-trained monocular visual odometry network, DeepVO, extracts camera pose information to guide the network in better analyzing temporal sequences.

In point-supervised training, only the video sequence feature encoder from the unsupervised pretext task is used for re-embedding raw features; the video-level feature encoder is not used. The weights of the video sequence feature encoder are obtained from the unsupervised pretext task, while other network weights are randomly initialized.

At the start of training, the training set is traversed to extract embedding features corresponding to the annotated defect time-points, which are clustered using cosine similarity as the distance metric to obtain a set of defect prototype features. Similarly, features are extracted at regular intervals from videos annotated as not containing pipe defects, and clustered to obtain a set of background prototype features. The number of defect and background prototype features is a hyperparameter, forming the multi-prototype features memory.

For a video sequence, the embedding features are first input into the video frame sequence encoder, resulting in the embedded features:
\begin{equation}
  F_{embed} = VideoFrameSeqEncoder_Q(F_{input})
\end{equation}
The embedded features and multi-prototype features memory are then input into the prototype-aware decoder, which first performs self-attention on the embedded features, then cross-attention where the sample sequence embedding features act as the Query, and the prototype features act as Key and Value. This guides the network in learning defect features from cross-video prototype features:
\begin{equation}
  F_{percept} = ProtoAwareDecoder(F_{embed}, F_{prototype})
\end{equation}
Subsequently, a multi-layer perceptron classifier predicts background class scores $S_{bkg}$ and defect class scores $S_{defect}$ at the video frame level:
\begin{equation}
  S_{bkg}, S_{defect} = MLPs(F_{percept})
\end{equation}
Additionally, visual odometry features are used to enhance feature representation through a visual odometry feature encoder. The frame-level dimensions are mapped to one dimension through a multi-layer perceptron and then processed with softmax to obtain visual odometry attention scores, which are multiplied with $S_{defect}$ to further refine defect class frame-level scores:
\begin{equation}
  S_{defect} = S_{defect} \times softmax(MLPs(Encoder_{VO}(F_{VO})))
\end{equation}
Finally, non-maximum suppression and other post-processing methods are used to obtain the defect time-intervals in the video. The instance-level fine-tuning network from the HR-Pro method~\citep{zhang2023hr} is employed to refine the predicted time-intervals.

\subsection{Loss Functions}

The loss function of PipeSPO consists of two major components: the unsupervised loss function and the point-supervised loss function. To facilitate the description, this section uses the following formula to compute the exponential scaling of the cosine similarity between two features $F_{1}$ and $F_{2}$:
\[
  S(F_{1}, F_{2}) = \exp \left( \frac{F_{1} \cdot F_{2}}{\|F_{1}\| \|F_{2}\|} \cdot \frac{1}{\tau} \right)
\]
where $\tau$ is a temperature hyperparameter used to control the scale of similarity scores, thus regulating the discrimination difficulty of features.

\subsubsection{Unsupervised loss}

The loss function for the unsupervised stage employs the InfoNCE loss, a contrastive loss designed to increase the similarity between the query feature and the positive sample feature while decreasing the similarity between the query feature and all negative sample features. Additionally, it also reduces the similarity among negative samples themselves. This design forces the network to learn the temporal information of video sequences, thereby enhancing its ability to locate time-intervals. The loss function is defined as follows:
\begin{equation}
  L_{NCE} = -\log \frac{S(F_{q}, F_{pos})}{S(F_{q}, F_{pos}) + \sum_{i=1}^{Q} S(F_{q}, F_{neg_i})}
\end{equation}
where $F_{q}$ and $F_{pos}$ are the query feature and the positive sample feature of the video, respectively, $F_{neg_i}$ is the feature of the $i$-th negative sample, and $Q$ is the size of the negative sample queue, representing the number of negative samples.

\subsubsection{Point-supervised loss}

The loss function in the point-supervised stage consists of two parts: the base loss, which is a focal loss based on temporal sequences, and a multi-prototype-based contrastive loss.

\textbf{Base Loss}: The base loss function is derived from the HR-Pro model~\citep{zhang2023hr}. It utilizes the characteristic that each pipe defect contains a time-point, and adjacent time-points belong to different pipe defects. Based on the time-points and defect scores, the total set of defect time-points and pseudo defect time-points, $T^+ = \{t_i\}_{i=1}^{N_{def}}$, and the set of pseudo background time-points, $T^- = \{t_j\}_{j=1}^{N_{bkg}}$, are constructed. Specifically, if the defect score of a frame near a defect time-point exceeds a given threshold, it is marked as a pseudo defect frame. Conversely, segments between two adjacent defect time-points with defect scores below the threshold are marked as pseudo background segments. The focal loss is calculated using these real and pseudo annotations, with two hyperparameters controlling the loss weights:
\begin{equation}
  L_{\text{focal}} = \lambda_{f_{d}} \frac{1}{N_{\text{d}}} \sum_{i=1}^{N_{\text{d}}} FL(S_{d}, t_i) + \lambda_{f_{b}} \frac{1}{N_{\text{b}}} \sum_{j=1}^{N_{\text{b}}} FL(1 - S_{d}, t_j)
\end{equation}
Here, $N_{d}$ is the total number of defect time-points, $N_{b}$ is the number of pseudo background frames, FL represents the focal loss function~\citep{lin2017focal}, $S_{d}$ is the frame-level defect score, $t_i$ is the time-point of the defect frame, and $t_j$ is the time-point of the pseudo background frame.

\textbf{Prototype Contrastive Loss}: The prototype contrastive loss is based on multi-prototype contrastive loss. After obtaining the total set of defect time-points and pseudo background time-points via frame-level defect scores, the features corresponding to the time-points in the embedding feature $F_{embed}$ are extracted to form the feature sets for defect time-points, $F_{d}$, and pseudo background time-points, $F_{b}$. The prototype contrastive loss consists of two parts: the defect frame prototype contrastive loss and the background frame prototype contrastive loss.

The defect time-point contrastive loss is calculated as follows:
\begin{equation}
  L_{def} = -\frac{1}{N_{d}} \sum_{i=1}^{N_{d}} \log \left( \frac{S(f_{d_i}, p_{d})}{S(f_{d_i}, p_{d}) + \sum_{j=1}^{N_{b}} S(f_{d_i}, p_{b_j})} \right)
\end{equation}
Here, \(N_{d}\) is the number of defect features, \(f_{d_i}\) is the \(i\)-th defect feature, \(p_{d}\) is the prototype feature most similar to \(f_{d_i}\), and \(p_{b_j}\) is the \(j\)-th background prototype feature.

The background time-point contrastive loss is calculated as follows:
\begin{equation}
  L_{bkg} = -\frac{1}{N_{b}} \sum_{k=1}^{N_{b}} \log \left( \frac{S(f_{b_k}, p_{b})}{S(f_{b_k}, p_{b}) + \sum_{l=1}^{N_{d}} S(f_{b_k}, p_{d_l})} \right)
\end{equation}
Similarly, in the formula, \(N_{b}\) is the number of background features, \(f_{b_k}\) is the \(k\)-th background feature, \(p_{b}\) is the prototype feature most similar to \(f_{b_k}\), and \(p_{d_l}\) is the \(l\)-th defect prototype feature.

These two formulas correspond to the contrastive loss calculations for defect time-points and background time-points, respectively. The objective is to maximize the similarity of each point to the most similar prototype of its respective category while minimizing the similarity to prototypes of other categories, thereby extending the supervision signal for time-point level annotations.

The prototype contrastive loss is the sum of the defect time-point contrastive loss and the background time-point contrastive loss, controlled by two hyperparameters:
\begin{equation}
  L_{proto} = \lambda_{p_d} L_{def} + \lambda_{p_b} L_{bkg}
\end{equation}

The total loss function in the point-supervised stage is the sum of the base loss and the prototype contrastive loss:
\begin{equation}
  L_{total} = L_{focal} + L_{proto}
\end{equation}

\section{Experiments and results}

\subsection{Technical details}\label{sec:pipespo_technical_details}

This study implements the PipeSPO model using the PyTorch framework~\citep{Paszke_PyTorch_An_Imperative_2019}, referring to the open-source code of the HR-Pro model~\citep{zhang2023hr}. All experiments were conducted on an Nvidia RTX 4090. The hyperparameter settings are as follows: for the unsupervised pretext task, the SGD optimizer~\citep{robbins1951stochastic} is used with a learning rate of 0.06, momentum of 0.95, weight decay of 0.02, temperature parameter of 0.01, and a negative sample queue size of 2200. For the point-supervised training, the AdamW optimizer~\citep{loshchilov2017decoupled} is used with a learning rate of 0.0001, weight decay of 0.001, contrastive loss temperature parameter of 0.85, $\lambda_{f_d}=0.6$, $\lambda_{f_b}=0.4$, $\lambda_{p_d}=0.3$, and $\lambda_{p_b}=0.75$. The evaluation metric is Average Precision (AP) at different Intersection over Union (IoU) thresholds, denoted as the average AP score.

This study uses a self-trained pipe image classification network as the baseline model. By setting different defect confidence thresholds, frames exceeding the threshold are identified, converting the predicted frame-level defect confidence scores into time-intervals as a post-processing step, thus obtaining predictions for different defect confidence thresholds.

This study compares the model with state-of-the-art methods in video summarization and point-supervised temporal action localization tasks. For the state-of-the-art PGL-SUM model~\citep{pgl_sum_2021} in video summarization, which requires time-interval level annotations, this study uses a pixel-based contrastive method to extend time-point level annotations. Since the video summarization post-processing step generates a fixed proportion of summaries for all videos, which is not suitable for this task, this study uses the frame importance scores output by the PGL-SUM model with the same post-processing step as the baseline method to evaluate performance on the temporal localization task. Due to their significantly worse performance compared to the baseline, other methods weaker than the PGL-SUM model in this task are not included in the comparison.

This study compares all point-supervised temporal action localization models, including the SF-Net model~\citep{ma2020sf}, LACP model~\citep{LACP_2021}, and the state-of-the-art HR-Pro model~\citep{zhang2023hr}. The weakest performing SF-Net model failed to converge on this dataset with both default and manually tuned hyperparameters, and thus is not included in the comparative results table.

\subsection{Comparative experiments}

\begin{table*}
  \centering
  \setlength{\tabcolsep}{10pt}
  \caption{Comparative experimental results of different models under different IoU thresholds.}
  \begin{threeparttable}
    \begin{tabular}{@{}l|ccccccc|ccc@{}}
      \toprule
      \multirow{2}{*}{Model} & \multicolumn{7}{c|}{AP@IoU (\%)} & \multicolumn{3}{c}{Average (\%)}                                                                                                                                        \\
                             & 0.1                              & 0.2                              & 0.3            & 0.4            & 0.5            & 0.6            & 0.7           & (0.1:0.5)      & (0.3:0.7)      & (0.1:0.7)      \\ \midrule
      Baseline               & 29.55                            & 19.78                            & 15.39          & 10.52          & 6.23           & 2.68           & 1.92          & 16.30          & 7.35           & 12.30          \\
      PGL-SUM\tnote{a}       & 13.05                            & 9.10                             & 6.21           & 4.58           & 2.15           & 0.90           & 0.52          & 7.02           & 2.87           & 5.22           \\
      LACP\tnote{b}          & 49.59                            & 34.49                            & 26.50          & 20.87          & 13.78          & 7.77           & 3.51          & 29.04          & 14.49          & 22.36          \\
      HR-Pro\tnote{c}        & 65.10                            & 52.32                            & 42.60          & 34.28          & 22.76          & 11.54          & 7.65          & 43.41          & 23.77          & 33.75          \\
      \textbf{PipeSPO}       & \textbf{74.50}                   & \textbf{65.90}                   & \textbf{52.30} & \textbf{42.40} & \textbf{30.20} & \textbf{19.60} & \textbf{8.20} & \textbf{53.09} & \textbf{30.56} & \textbf{41.89} \\ \bottomrule
    \end{tabular}
    \begin{tablenotes}
      \footnotesize
      \item[a] From~\cite{pgl_sum_2021}
      \item[b] From~\cite{LACP_2021}
      \item[c] From~\cite{zhang2023hr}
    \end{tablenotes}
  \end{threeparttable}
  \label{tab:pipespo_comp_results}
\end{table*}

The comparative experiments encompass three types of methods. First, the baseline method utilizes frame-by-frame classification via an image network. Subsequently, the PGL-SUM model from the video summarization, along with the LACP and HR-Pro models from the point-supervised TAL, are considered, representing the leading techniques in their respective tasks. Technical details can be found in Section~\ref{sec:pipespo_technical_details}. All results are evaluated on the same test set, with comparative methods employing default hyperparameters and optimal hyperparameter combinations identified through grid search during post-processing. Experimental results are shown in Table~\ref{tab:pipespo_comp_results}. It is observed that PipeSPO achieves superior performance across all IoU thresholds, with an average precision of 41.89\% over 0.1-0.7 IoU thresholds, marking an improvement of 8.14\% over the state-of-the-art HR-Pro model.

The baseline method achieves an average precision of 12.30\% over 0.1-0.7 IoU thresholds. The subpar performance of the baseline method is attributed to its inability to analyze temporal sequences, as it relies on frame-by-frame classification via an image network. Furthermore, the accuracy of the pipe image classification network, trained with time-point level annotations, does not meet the requirements for time-interval localization.

The state-of-the-art PGL-SUM model in the video summarization achieves an average precision of 5.22\% over 0.1-0.7 IoU thresholds. The performance of the PGL-SUM model is limited by its requirement for time-interval level annotations. This study employs pixel-based contrastive methods and minimum interval length settings to convert time-interval level annotations to frame-level pseudo annotations. However, these pseudo annotations lack sufficient accuracy, resulting in the PGL-SUM model performing comparatively worse than the baseline method.

For the LACP and HR-Pro models, which performed much better in the point-supervised TAL, significant improvement over the baseline method is observed. Specifically, the LACP model achieves an average precision of 22.36\% over 0.1-0.7 IoU thresholds, while the HR-Pro model surpasses the LACP model, achieving an average precision of 33.75\% over 0.1-0.7 IoU thresholds. Both models utilize image classification features and optical flow features to capture color and motion information in video sequences. The LACP model dynamically generates pseudo labels using a greedy algorithm at the end of each training epoch based on intermediate model outputs to enhance performance under point-supervision. The HR-Pro model employs cross-video defect category prototypes and instance-level fine-tuning networks to achieve better localization. However, these networks are designed for temporal action localization tasks and do not account for the specific characteristics of pipe defect temporal localization tasks, resulting in certain limitations.

The proposed PipeSPO model achieves the highest performance, with an average precision of 41.89\% over 0.1-0.7 IoU thresholds. Compared to the state-of-the-art HR-Pro model, PipeSPO surpasses existing frame-by-frame detection methods based on image classification networks, with an average precision improvement of 9.68\% over 0.1-0.5 IoU thresholds, 6.79\% over 0.3-0.7 IoU thresholds, and 8.14\% over 0.1-0.7 IoU thresholds, demonstrating significant enhancement. This improvement is attributed to PipeSPO addressing the shortcomings of existing methods in the pipe context by utilizing unsupervised contrastive learning as a pretext task to enhance the representation capability of the video sequence encoder. It extracts prototype features for defect time-points and background time-points through clustering-based methods and employs prototype contrastive loss to obtain additional supervision signals. Additionally, it uses a pre-trained monocular visual odometry network, DeepVO, on the KITTI dataset to extract camera pose information, guiding the network to better analyze temporal sequences. Finally, it replaces optical flow features with 3D-DWT features, which are more suitable for pipe scenarios, to achieve better performance.

\subsection{Ablation study}

To validate the effectiveness of the PipeSPO model, a series of ablation experiments were conducted. These experiments included tests on the dynamic information feature extractor, the unsupervised pretext task, the clustering-based prototype-aware decoder, the prototype contrastive loss, the visual odometry attention module, and the position encoding. In the ablation experiments, only one hyperparameter was altered at a time, with all other parameters kept constant.

\subsubsection{Dynamic information feature extractor}

\begin{table}
  \centering
  \setlength{\tabcolsep}{1pt}
  \caption{Ablation study results on the dynamic information feature extractor in the PipeSPO model. PipeSPO by default uses the \textit{3D-DWT} extractor, and \textit{None} indicates no dynamic information features are used.}
  \begin{tabular}{@{}l|cccccc@{}}
    \toprule
    \multirow{2}{*}{Name} & \multicolumn{6}{c}{Average AP@IoU (\%)}                                                                                                                          \\
                          & (0.1:0.5)                               & \multicolumn{1}{c|}{\( \Delta \)} & (0.3:0.7)      & \multicolumn{1}{c|}{\( \Delta \)} & (0.1:0.7)      & \( \Delta \) \\ \midrule
    None                  & 30.27                                   & \multicolumn{1}{c|}{-22.82}       & 11.47          & \multicolumn{1}{c|}{-19.09}       & 22.83          & -19.06       \\
    Optical Flow          & 43.15                                   & \multicolumn{1}{c|}{-9.94}        & 20.87          & \multicolumn{1}{c|}{-9.69}        & 33.02          & -8.88        \\
    3D-DWT                & \textbf{53.09}                          & \multicolumn{1}{c|}{-}            & \textbf{30.56} & \multicolumn{1}{c|}{-}            & \textbf{41.89} & -            \\ \bottomrule
  \end{tabular}
  \label{tab:pipespo_abl_feature_extractor_motion}
\end{table}

As shown in Table~\ref{tab:pipespo_abl_feature_extractor_motion}, the model performs best when using the 3D-DWT feature extractor. Compared to the scenario where no dynamic information features are used, a 19.06\% performance improvement is achieved at 0.1-0.7 IoU thresholds. The performance of 3D-DWT features surpasses that of optical flow features, with an 8.88\% higher performance at 0.1-0.7 IoU thresholds, demonstrating the effectiveness of 3D-DWT features in pipe scenarios. Since dynamic information in videos contains rich temporal information about pipe defects, the dynamic information feature extractor is crucial for the point-supervised pipe defect temporal localization task.

\subsubsection{Unsupervised pretext task}

\begin{table}
  \centering
  \setlength{\tabcolsep}{1.5pt}
  \caption{Ablation study results on the unsupervised pretext task in the PipeSPO model.}
  \begin{threeparttable}
    \begin{tabular}{@{}l|cccccc@{}}
      \toprule
      \multirow{2}{*}{Name}                        & \multicolumn{6}{c}{Average AP@IoU (\%)}                                                                                                                          \\
                                                   & (0.1:0.5)                               & \multicolumn{1}{c|}{\( \Delta \)} & (0.3:0.7)      & \multicolumn{1}{c|}{\( \Delta \)} & (0.1:0.7)      & \( \Delta \) \\ \midrule
      No Unsup\tnote{$\dagger$}\phantom{$\dagger$} & 41.06                                   & \multicolumn{1}{c|}{-12.03}       & 20.02          & \multicolumn{1}{c|}{-10.54}       & 31.24          & -10.65       \\
      Unsup\tnote{$\ddagger$}\phantom{$\ddagger$}  & \textbf{53.09}                          & \multicolumn{1}{c|}{-}            & \textbf{30.56} & \multicolumn{1}{c|}{-}            & \textbf{41.89} & -            \\ \bottomrule
    \end{tabular}
    \begin{tablenotes}
      \footnotesize
      \item[$\dagger$] No unsupervised pretext task
      \item[$\ddagger$] With unsupervised pretext task
    \end{tablenotes}
  \end{threeparttable}
  \label{tab:pipespo_abl_unsupervised}
\end{table}

As shown in Table~\ref{tab:pipespo_abl_unsupervised}, without the unsupervised pretext task, the model's performance significantly drops, with average AP scores decreasing by 12.03\%, 10.54\%, and 10.65\% for the different thresholds, respectively. The performance drop can be attributed to several reasons. First, the unsupervised pretext task leverages a large amount of unlabeled data to enhance the representation capability of the video sequence encoder, allowing the network to better analyze temporal sequences. Second, PipeSPO uses features from multiple clusters for background and prototype classes. With the unsupervised pretext task, the video sequence feature encoder is well-trained, enabling it to better integrate image classification features, 3D-DWT features, and temporal sequence information. Consequently, when extracting prototypes across videos at the start of point-supervised training, the prototypes have better representation capabilities compared to using randomly initialized weights, thereby improving the model's performance during point-supervised training.

\subsubsection{Multi-prototype features}

\begin{table*}
  \centering
  \setlength{\tabcolsep}{14pt}
  \caption{Ablation study results on the prototype feature modules in the PipeSPO model. The first row represents not using any prototype feature-related modules.}
  \begin{threeparttable}
    \begin{tabular}{@{}p{50pt}<{\centering}c|c|c|cc@{}}
      \toprule
      \multicolumn{2}{c|}{Number of Prototype} & \multirow{2}{*}{Prototype-Aware Decoder} & \multirow{2}{*}{Prototype Contrastive Loss} & \multicolumn{2}{c}{Average AP@IoU (\%)}                                 \\
      N\tnote{$\dagger$}                       & 1                                        &                                             &                                         & (0.1:0.7)      & \( \Delta \) \\ \midrule
                                               &                                          &                                             &                                         & 20.84          & -21.05       \\ \midrule
      \checkmark                               &                                          &                                             & \checkmark                              & 24.08          & -17.81       \\
      \checkmark                               &                                          & \checkmark                                  &                                         & 33.31          & -8.58        \\ \midrule
                                               & \checkmark                               & \checkmark                                  & \checkmark                              & 40.37          & -1.52        \\
      \checkmark                               &                                          & \checkmark                                  & \checkmark                              & \textbf{41.89} & -            \\ \bottomrule
    \end{tabular}
    \begin{tablenotes}
      \footnotesize
      \item[$\dagger$] \textit{N} here is a hyperparameter greater than 1, representing the number of prototype features.
    \end{tablenotes}
  \end{threeparttable}
  \label{tab:pipespo_abl_proto}
\end{table*}

As shown in the ablation study results in Table~\ref{tab:pipespo_abl_proto}, the PipeSPO model performs best when using multi-prototype features, the prototype-aware decoder, and the prototype contrastive loss. Replacing multi-prototype features with single-prototype features decreases the model performance by 1.52\%. The number of prototype features in PipeSPO is determined based on the dataset size and the feature extraction capability of the feature extractor. Multi-prototype features can better adapt to feature extractors with weaker extraction capabilities, thereby improving model performance.

When the prototype contrastive loss is omitted and only the prototype-aware decoder is used, the model performance decreases by 8.58\%. This indicates that the cross-attention operation between video sequence embedding features and prototype features helps the network better focus on potential typical defect features and background features in the video sequence, thereby enhancing model performance. Conversely, when the prototype-aware decoder is omitted and only the prototype contrastive loss is used, the model performance decreases by 17.81\%. This demonstrates that the prototype-aware decoder has a significant impact on model performance as the network cannot directly perceive the prototype features during training without it. It can only indirectly utilize the information of prototype features through the loss value during backpropagation, leading to performance degradation. When neither of the prototype feature-related modules is used, the model performs the worst, with a performance drop of 12.47\% compared to using only the prototype-aware decoder and a drop of 3.24\% compared to using only the prototype contrastive loss. This also shows that directly utilizing prototype features in the prototype-aware decoder has a greater impact on model performance enhancement than indirectly utilizing prototype features in the prototype contrastive loss.

\subsubsection{Visual odometry attention module}

\begin{table}
  \centering
  \setlength{\tabcolsep}{2pt}
  \caption{Ablation study results on the visual odometry attention module.}
  \begin{threeparttable}

    \begin{tabular}{@{}l|cccccc@{}}
      \toprule
      \multirow{2}{*}{Name}                     & \multicolumn{6}{c}{Average AP@IoU (\%)}                                                                                                                          \\
                                                & (0.1:0.5)                               & \multicolumn{1}{c|}{\( \Delta \)} & (0.3:0.7)      & \multicolumn{1}{c|}{\( \Delta \)} & (0.1:0.7)      & \( \Delta \) \\ \midrule
      No VO\tnote{$\dagger$}\phantom{$\dagger$} & 40.74                                   & \multicolumn{1}{c|}{-12.35}       & 17.08          & \multicolumn{1}{c|}{-13.48}       & 30.55          & -11.34       \\
      VO\tnote{$\ddagger$}\phantom{$\ddagger$}  & \textbf{53.09}                          & \multicolumn{1}{c|}{-}            & \textbf{30.56} & \multicolumn{1}{c|}{-}            & \textbf{41.89} & -            \\ \bottomrule
    \end{tabular}
    \begin{tablenotes}
      \footnotesize
      \item[$\dagger$] No visual odometry attention module
      \item[$\ddagger$] With visual odometry attention module
    \end{tablenotes}
  \end{threeparttable}
  \label{tab:pipespo_abl_vo}
\end{table}

As shown in the ablation study results in Table~\ref{tab:pipespo_abl_vo}, the model performance drops significantly when the visual odometry attention module is not used. This reduction occurs because the visual odometry attention module extracts camera pose information to calculate visual odometry attention scores. Due to the unique characteristics of pipe inspection videos, the probability of pipe defects appearing increases when the inspection vehicle stops and rotates the camera. By extracting camera pose information through a pre-trained monocular visual odometry network, this module leverages this characteristic by multiplying the visual odometry attention scores with the network's predicted defect class scores, thereby increasing the defect class scores where defects are more likely to appear. This approach enables the network to use camera pose information for better predictions.

\subsubsection{Positional encoding}

\begin{table}
  \centering
  \setlength{\tabcolsep}{2pt}
  \caption{Ablation study results on positional encoding in the PipeSPO model.}
  \begin{threeparttable}
    \begin{tabular}{@{}l|cccccc@{}}
      \toprule
      \multirow{2}{*}{Name}                     & \multicolumn{6}{c}{Average AP@IoU (\%)}                                                                                                                          \\
                                                & (0.1:0.5)                               & \multicolumn{1}{c|}{\( \Delta \)} & (0.3:0.7)      & \multicolumn{1}{c|}{\( \Delta \)} & (0.1:0.7)      & \( \Delta \) \\ \midrule
      No PE\tnote{$\dagger$}\phantom{$\dagger$} & 16.67                                   & \multicolumn{1}{c|}{-36.42}       & 5.16           & \multicolumn{1}{c|}{-25.40}       & 12.50          & -29.39       \\
      PE\tnote{$\ddagger$}\phantom{$\ddagger$}  & \textbf{53.09}                          & \multicolumn{1}{c|}{-}            & \textbf{30.56} & \multicolumn{1}{c|}{-}            & \textbf{41.89} & -            \\ \bottomrule
    \end{tabular}
    \begin{tablenotes}
      \footnotesize
      \item[$\dagger$] No positional encoding
      \item[$\ddagger$] With positional encoding
    \end{tablenotes}
  \end{threeparttable}
  \label{tab:pipespo_abl_positional_encoding}
\end{table}

As shown in the ablation study results in Table~\ref{tab:pipespo_abl_positional_encoding}, the model performance decreases significantly when positional encoding is not used. This decline occurs because the encoder module of the transformer architecture cannot perceive sequence order, resulting in the video-level features encoded by the video frame sequence encoder during pretext task training only representing the presence of pipe defects within the detection video but not their temporal location. In the point-supervised stage, the lack of positional information similarly weakens the network's ability to analyze temporal sequences, leading to a significant drop in model performance.

\section{Conclusion and future work}

This study proposes a semi-supervised model named PipeSPO for the temporal localization of defects in drainage pipe inspection videos. By employing an unsupervised contrastive learning pretext task, the video frame sequence encoder is trained without annotations, significantly alleviating the issue of weak feature representation in pipe scenarios. This establishes a foundation for extracting defect prototypes and background prototypes under weak supervision. Using a clustering-based method, numerous defect and background time-point features in the dataset are clustered to obtain prototype feature sets for defect and background time-points. During training, pseudo defect and pseudo background time-points are dynamically selected, and a prototype contrastive loss function is designed to provide the network with additional supervision signals. Additionally, the monocular visual odometry network DeepVO, pre-trained on the KITTI dataset, is used to extract camera pose information, guiding the network to better analyze the temporal sequence. Experimental results demonstrate that PipeSPO achieves superior performance on real-world pipe inspection video datasets.

However, PipeSPO still has some limitations, such as the inclusion of complex post-processing steps and the inability to predict the number, category, and severity of pipe defects in a given time-interval. Future work could focus on multitask learning and simplifying model complexity to comprehensively enhance practicality and performance.

\bibliography{wileyNJD-APA}

\end{document}